\documentclass[10pt,a4paper]{article}
\usepackage{geometry}

\usepackage[utf8]{inputenc}
\usepackage[greek,english]{babel}

\usepackage{lmodern}
\usepackage{listings}
\usepackage{pgfplotstable}
\usepackage{cite}
\usepackage{hyperref}
\usepackage{authblk}
\usepackage{graphicx}
\usepackage{caption}
\usepackage{algorithm}
\usepackage{algorithmicx}
\usepackage{algpseudocode}

\newcommand{\comment}[1]{}
\newcommand{\todo}[1]{{\color{red}#1}}
\newcommand{\new}[1]{#1}

\newcommand{\projectDescription}[5]{
\begin{table}[H]
	\centering
	\begin{tabular}{|p{20mm}|p{115mm}|}\hline
		Project Id 						&  #1 \\\hline
		Process 						&  #2 \\\hline
		Components:  			&  #3 \\\hline
		Time spent:  					&  #4 \\\hline
		Accuracy 						&  #5 \\\hline
	\end{tabular}
\end{table}
}

\title{CS563-QA: A  Collection for Evaluating \\Question Answering Systems
    }
\author[1,2]{Katerina Papantoniou}
\author[1,2]{Yannis Tzitzikas}
\affil[1]{Computer Science Department, University of Crete, Greece}
\affil[2]{Institute of Computer Science, FORTH-ICS, Greece}
\affil[ ]{\textit {\{kpapantoniou,tzitzik\}@csd.uoc.gr}}
\usepackage{color}
\usepackage{paralist, tabularx}

\begin{document}
\maketitle

\comment{
\title{CS563-QA: A  Collection for Evaluating \\Question Answering Systems\\
            \todo{(Draft 0.3)}\footnote{\todo{Some parts of this report are not yet complete}}
      }
\author{Katerina Papantoniou and Yannis Tzitzikas}

\address{Computer Science Department, University of Crete, and \\
        Institute of Computer Science, 
        Foundation for Research and Technology - Hellas (FORTH) \\
        email:  tzitzik, kpapantoniou@csd.uoc.gr}

==}
\begin{abstract}
Question Answering (QA) is a challenging topic since it requires tackling the various difficulties of natural language understanding.
Since evaluation is important not only for identifying the strong and weak points of the various techniques for QA,  
but also for facilitating the inception of new methods and techniques, 
in this paper  we present a collection for evaluating QA methods over free text
that we have created.  
Although it is a small collection, it contains cases of increasing difficulty, 
therefore it has an educational value and it can be  used for rapid evaluation of QA systems.
\end{abstract}


\section{Introduction}
\label{S:1}

Question Answering (QA) systems aim at providing precise answers to questions, 
posed by users using natural language. 
Such systems are used in a wide range of application areas.  
QA is a challenging task since it requires tackling the various difficulties of natural language. 
Although the first QA systems were created long ago (back in 1960s), the problem is still open,
the existing techniques have several limitations \cite{rodrigo2017study},
and therefore QA is subject of continuous research.
There is a wide range of techniques for QA ranging from simple regular expression-based methods, 
to methods relying on deep learning,
and there are several survey papers including \cite{wang2006survey,mishra2016survey,DBLP:journals/corr/Patra17}.

Since {\em evaluation} is important not only for identifying the strong and weak points of the various techniques 
(as well as their prerequisites),  
but also for facilitating the inception of new methods and techniques, 
in this paper we present a collection for evaluating QA methods. 
We focus on QA over plain text, i.e. the collection comprises free text.

This collection was constructed in the context of the graduate course CS563 of the 
Computer Science Department of the University of Crete (Spring 2019).

The rest of this paper is organized as follows
Section \ref{sec:Objectives} describes the objectives,
Section \ref{sec:TheCollection} describes the collection,
Section \ref{sec:EvaluationMetrics} briefly discusses metrics that can be used,
Section \ref{sec:BestProject} describes how the collection was used in the course
and what the students achieved,
and finally,
Section \ref{sec:HowToGet} provides information about how to get the collection.

\section{Objectives}
\label{sec:Objectives}

Although there are several collections for evaluating QA systems (see \cite{QASurvey2019Ongoing}),
the current  collection has been  constructed based on the following objectives:
\begin{compactitem}
\item   It should be small so that one can run experiments very {\em fast}.
\item 	It should contain cases that require tackling {\em various kinds of difficulties}.
\item 	It should contain cases of {\em increasing difficulty}.
\item 	It should have {\em educational value}.
\end{compactitem}

\section{The Collection}
\label{sec:TheCollection}

The collection consists of {\em topics}.
Each topic has an id, a title, a text 
(ranging from 40 to 18310 words), 
and a list of
question-answer pairs.
On average each topic has 5 question-answer pairs.
The collection is represented in JSON (JavaScript Object Notation) format, 
an  example
is shown in 
Figure \ref{fig:example}
in Appendix.
In total the collection contains 149 question-answer pairs.

The topics are organized in groups of different levels of difficulty, starting from a group of relatively easy questions, 
ending up to a group with very hard to answer questions, or even impossible for the state-of-the-art methods.
The groups are described in Table \ref{tbl:groups}.

\begin{table}[h]
	\centering
\begin{tabular}{l p{4cm} p{65mm}}
\hline
\textbf{Group of Topics} & \textbf{Difficulty} & \textbf{Indicative Tools/Resources}\\
\hline
 1-12       & Can be answered easily 
 				& Stanford CoreNLP \\\hline
12-24      & Lexical gap  
 				& Semantic dictionaries and ontologies, word embeddings (e.g. WordNet, Wiktionary, pretrained word embeddings) \\\hline
 25-41      & Disambiguation, Inferences 	& world and commonsense knowledge (e.g.  DBPedia Spotlight, DBPedia Ontology, YAGO )   \\\hline
\end{tabular}
\caption{Groups of topics}
\label{tbl:groups}
\end{table}

\subsection{The First Group}
\label{sec:GroupA}

The {\em first group} (containing topics 1 to 12) consists of questions that do not require complex natural language processing to find the answers. 
The extraction of the answer is quite straightforward by applying basic linguistic analysis and by searching for simple patterns in the given snippets. 
This group includes questions that require
tackling some form of noise (e.g.  spelling errors) and possible the use of gazetteers.
For example,  indicative text-query-answer combinations  of this category are
shown in Table \ref{tbl:groupA}.

\begin{table}[htbp]
\centering
\begin{tabular}{|l|p{12cm}|}\hline
Topic & Wolfgang Amadeus Mozart \\\hline
Text				&  Wolfgang Amadeus Mozart(27 January 1756 – 5 December 1791), to Leopold Mozart (1719 –1789) and Anna Maria, nee Pertl (1720 – 1778), was a prolific and influential composer of the classical era.
Born in Salzburg, Mozart showed prodigious ability from his earliest childhood. \\\hline
Question1: 		& Where was Mozart born? \\\hline
Answer1: 		& Salzburg \\\hline
Question3: 		& What was the first name of Mozart's father?\\\hline
Answer3: 		& Leopold \\\hline
Question4: 		& What was Mozart's profession?\\\hline
Answer4: 		& composer \\\hline
\end{tabular}
\caption{Examples from Group A}
\label{tbl:groupA}
\end{table}

\subsection{The Second Group}
\label{sec:GroupB}

The {\em second group}  of questions is composed by questions that require more complex linguistic manipulations in order to extract the correct answer. 
In this category, a lexical gap may be exist between the keywords in question  and the content of the given snippet. 
Moreover, ambiguities may exist, e.g. POS ambiguity.
The recognition of semantic relations is required in this group of questions that can be extracted with the help of semantic dictionaries, ontologies,  word embeddings, etc.
For example,  indicative text-query-answer combinations  of this category are
shown in Table \ref{tbl:groupB}.

\begin{table}[htbp]
\centering
\begin{tabular}{|l|p{12cm}|}\hline
Topic & Santiago de Compostela \\\hline
Text				& 
{\small
Santiago de Compostela is the capital of the autonomous community of Galicia, in northwestern Spain.
The city has its origin in the shrine of Saint James the Great, now the Cathedral of Santiago de Compostela, as the destination of the Way of St. James, a leading Catholic pilgrimage route since the 9th century. In 1985, the city's Old Town was designated a UNESCO World Heritage Site. The population of the city in 2012 was 95,671 inhabitants, while the metropolitan area reaches 178,695. In 2010 there were 4,111 foreigners living in the city, representing 4.3\% of the total population. The main nationalities are Brazilians (11\%), Portuguese (8\%) and Colombians (7\%).

By language, according to 2008 data, 21.17\% of the population always speak in Galician, 15\% always speak in Spanish, 31\% mostly in Galician and the 32.17\% mostly in Spanish. According to a Xunta de Galicia 2010 study the 38.\% of the city primary and secondary education students had Galician as their mother tongue. 
}\\\hline
Question1: 		& How many people live in the city?\\\hline
Answer1: 		& 95,671\\\hline
Question2: 		& What is the number of non-native Galician residents in Santiago de Compostela?  \\\hline
Answer2: 		& 4,111\\\hline 
\end{tabular}
\caption{Examples from Group B}
\label{tbl:groupB}
\end{table}

\subsection{The Third Group}
\label{sec:GroupC}

The {\em last} and more complex group of questions consists of questions that require 
(a) some sort of ambiguity to be resolved, 
(b) world and commonsense knowledge, e.g. inferences that are based on knowledge found on resources beyond the given snippet or knowledge that all humans are expected to know. 
Cases of semantic and syntactic ambiguity, erroneous, partial or implied information and cases that refer not only to objective facts but also on sentimental opinions fall in this category.
For these cases, a combination of the linguistic resources and tools of the previous category with world knowledge is required. 
Linguistic data and tools for the  Linked Open Data Cloud such as  DBPedia Ontology, YAGO and DBPedia Spotlight can be helpful is this group of questions.
For example,  indicative text-query-answer combinations  of this category are
shown in Table \ref{tbl:groupC} and \ref{tbl:groupC2}.

\begin{table}[htbp]
\centering
\begin{tabular}{|l|p{12cm}|}\hline
Text				&   
{\small
The Low Countries, the Low Lands, is a coastal lowland region in northwestern Europe, forming the lower basin of the Rhine, Meuse, and Scheldt rivers, divided in the Middle Ages into numerous semi-independent principalities that consolidated in the countries of Belgium, Luxembourg, and the Netherlands, as well as today's French Flanders. Historically, the regions without access to the sea have linked themselves politically and economically to those with access to form various unions of ports and hinterland, stretching inland as far as parts of the German Rhineland. It is why that nowadays some parts of the Low Countries are actually hilly, like Luxembourg and the south of Belgium. Within the European Union the region's political grouping is still referred to as the Benelux (short for Belgium-Netherlands-Luxembourg). 
}\\\hline
Question1: 		& What countries belong to the Netherlands? \\\hline
Answer1: 		& Belgium, Luxembourg, the Netherlands \\\hline
Question2: 		& In which way Benelux countries are linked?  \\\hline
Answer2: 		& politically and economically \\\hline
\end{tabular}
\caption{Examples from Group C}
\label{tbl:groupC}
\end{table}

\begin{table}[htbp]
\centering
\begin{tabular}{|l|p{12cm}|}\hline
Text				&    As a freshman, he was a member of the Tar Heels' national championship team in 1982. Jordan joined the Bulls in 1984 as the third overall draft pick. He quickly emerged as a league star and entertained crowds with his prolific scoring. His leaping ability, demonstrated by performing slam dunks from the free throw line in Slam Dunk Contests, earned him the nicknames Air Jordan and His Airness. He also gained a reputation for being one of the best defensive players in basketball. In 1991, he won his first NBA championship with the Bulls, and followed that achievement with titles in 1992 and 1993, securing a \"three-peat\". Although Jordan abruptly retired from basketball before the beginning of the 1993–94 NBA season, and started a new career in Minor League Baseball, he returned to the Bulls in March 1995 and led them to three additional championships in 1996, 1997, and 1998, as well as a then-record 72 regular-season wins in the 1995–96 NBA season. Jordan retired for a second time in January 1999, but returned for two more NBA seasons from 2001 to 2003 as a member of the Wizards.
\\\hline
Question1: 		& What team did Michael Jordan play for \textbf{after} the Bulls? \\\hline
Answer1: 		& Washington Wizards/Wizards \\\hline
\end{tabular}
\caption{Examples from Group C-temporal}
\label{tbl:groupC2}
\end{table}

Finally, this category also includes questions that require analyzing the pragmatics of the text  
for answering the questions (co-reference resolution,  connections of sentences, speech acts, scripts, etc).
For example,  indicative text-query-answer combinations  of this category are
shown in Table \ref{tbl:groupC3}

\begin{table}[htbp]
\centering
\begin{tabular}{|l|p{12cm}|}\hline
Text			&   John likes Mary. He gives her a present.  \\\hline
Question1: 		&  Who likes Mary? \\\hline
Answer1: 		& John \\\hline
Question2: 		& What Mary received? \\\hline
Answer2: 		& A present \\\hline
\end{tabular}

\begin{tabular}{|l|p{12cm}|}\hline
Text				&   John had a meeting in the university. However, he left for the university quite late this morning and he missed the bus of 8:30. When he arrived, Tom met him in the corridor, just a few meters from the entrance of the meeting room. He told him that everyone was waiting for him. He took a big breath and entered into the room with a big smile..  \\\hline
Question1: 		&  Who smiled? \\\hline
Answer1: 		& John \\\hline
Question2: 		& How many were in the meeting room? \\\hline
Answer2: 		& At least two \\\hline
Question3: 		& What time John arrived at the meeting?\\\hline
Answer3: 		& After 8:30 \\\hline
\end{tabular}


\begin{tabular}{|l|p{12cm}|}\hline
Text				&    John scored two goals in the last 5 minutes of the game giving 3 points to his football team. 
                    The goalkeeper of the other team was very sad.   \\\hline
Question1: 		&  Why was the goalkeeper very sad? \\\hline
Answer1: 		& Because he got 2 goals over the last 5 minutes of the game and his team lost. \\\hline

\end{tabular}
\caption{More Examples from Group C}
\label{tbl:groupC3}
\end{table}


\comment{
    Overall, the topics cover various areas.
    Some indicative question-answer pairs from some topics,
    and a part of their textual description,
    are shown in Table \ref{tbl:topicsDetailed}
}






\clearpage
\subsection{What Phenomena are Covered by the Collection}
\label{sec:Coverage}

\noindent{\bf Answer Types.}
The correct answers have been tagged with an {\em entity type},
i.e. it is the answer type in the context of question answering.
The types of the  answers of the questions in the collection are several,
below we show them organized  categories:

\begin{compactitem}
\item {\em Time-related}: 
        Year, Date, Time, Duration, Hour-related.
\item {\em Agents}:
        Person, Organization, Team, Theatre, Company.
\item {\em Roles}:
        Profession.
\item {\em Locations}:        
    Location, Location name, Region.
\item {\em Artifacts, Materials, Ingredients}:
    Material, Ingredient, Object.
\item {\em Quantitative}:
        Number, Percentage, Population, List.
\item {\em Contact details}:
        Telephone, Address.
\item {\em Monetary}:
        Money.
\item {\em Terminology}:
        Medical Term,  Genre.
\item {\em Processes}:
        Recipe.
\item {\em Misc}:
        Aspect, URL.
\end{compactitem}

\ \\
\noindent{\bf Question Types.}
The collection tries to cover a large number of question types to 
cover a wide range of users information needs.
We use the {\em questions types} mentioned in \cite{mishra2016survey}.
Table \ref{tbl:questionTypeCoverage} shows
for each question type,
one or more indicative topics 
of the  collections 
that contain such a question
(the list is not exhausting) 
and how many (approximately) such questions exist  in the collection.




\begin{table}[!ht!]
\centering
\begin{tabular}{|l|l|l|}
\hline
Question Type   &  Indicative Topics &  Num of questions \\\hline\hline
Factoid         &   1,2,3, ...       &  $>$ 60 \\\hline
Confirmation    &   3, 22, 27, ...   &   $>$ 6   \\\hline
Definition      &   12, 28, ...      &  $>$ 4\\\hline
Causal          &   36               &  1 \\\hline
List            &   25, 32, 37       &  3 \\\hline
Opinionated     &   33               &  2 \\\hline
With Examples   &   37               &  2 \\\hline
Procedural      &   28               &  1 \\\hline
Comparative     &   36               &  2 \\\hline

\end{tabular}
\caption{Coverage of Questions Types}
\label{tbl:questionTypeCoverage}
\end{table}

\noindent{\bf Kinds of Difficulties.}
Moreover,
Table \ref{tbl:DifficultiesCoverage}
lists some phenomena/difficulties
and 
some indicative topics
of the  collections 
where each phenomenon
occurs
and how many such questions (at minimum) exist in the collection.


\begin{table}[!ht!]
\centering
\begin{tabular}{|l|l|l|}
\hline
Difficulty   &  Topics &  Num. of questions (at least) \\\hline\hline
morphological differences       &    13,14,16,17,...      & 32 \\\hline
wrong spelling of name          &   3, 36      &  2 \\\hline
syntax ambiguity                &   27, 29, 40      &  4 \\\hline
ambiguity of references         &  34, 35, 36    & 3\\\hline
WSD &  41, 42, 26 & 3\\\hline
temporal reasoning              &  25,35,37,38  & 6 \\\hline
spatial reasoning               &  26         & 2 \\\hline
opinion sentiment reasoning     &  33, 36 & 5\\\hline
comparison                      &  36, 39 & 3\\\hline
assumed script                  &  40         & 2\\\hline
assumed domain model            &  40         & 1\\\hline
general historical knowledge    &  25, 37   &  4 \\\hline
speech act identification       &   -        & 0 \\\hline
\end{tabular}
\caption{Coverage of Various Difficulties}
\label{tbl:DifficultiesCoverage}
\end{table}


\comment{
    
    \begin{table}[!ht!]
    \centering
    \begin{tabular}{|l|l|l|}
    \hline
    Type &  Answer & Description \\\hline
    Factoid & Abbreviation, Person, ... & require a single fact as answer \\\hline
    List &  & require multiple facts to be returned \\\hline
    Definition & & require a short textual description  \\\hline
    Confirmation & & yes/no  \\\hline
    Complex??? & &  textual phrase or list that it is not a fact \\\hline
    \hline
    \end{tabular}
    \caption{Questions Types}
    \label{tbl:questionTypes}
    \end{table}
==}


\section{Evaluation Metrics}
\label{sec:EvaluationMetrics}

We should note that creating the equivalent of a standard Information Retrieval test collection is a difficult problem. 
In a IR test collection, the unit that is judged, the document, has an unique identifier, and it is easy to decide whether a document retrieved is the same document that has been judged. For QA, the unit that is judged is the entire string that is returned by the system and quite often different QA runs return different answer strings, hence it  is difficult to determine automatically whether the difference between a new string and a judged string is significant with respect to the correctness of the answer. One method to tackle this problem is to use so-called answer patterns and accept a string as correct if it matches an answer pattern for the question, answer patterns are described in \cite{voorhees2000building}.
In general, there are several methods and metric to evaluate QA systems, e.g. see \cite{rodrigo2017study}.
Since we expect the answer of the system to be a single answer (not a ranked list of possible answers),
set-based metrics are appropriate (i.e. not metrics for ranked results).
Therefore for the collection at hand, 
we propose evaluating QA systems according to {\em accuracy}.
If $Q$ denotes the  set of questions that are used in the evaluation,
and $AQC$ those that were answered correctly,
then the Accuracy 
is the fraction of the questions
that were answered correctly 	i.e. $Accuracy =   \frac{|AQC|}{|Q|}$.

\subsection{Disclaimer}

Since the collection is small, any positive result
cannot be straightforwardly generalized. 
This collection is useful as a first test in the sense 
that if a QA process does not behave well in this collection, 
then certainly it will not work in a bigger collection or in a real world case.
If  a QA process behaves well in this collection, 
then this  does not necessarily mean that it will behave 
well in a bigger collection.

%

\section{Experience in the Course}
\label{sec:BestProject}

In the context of the course project of CS563,  
\new{
the Spring 2019 and Spring 2020 semesters, 
}
the students were given this collection and were asked
to build a QA system.
They were free to use whatever method, tool and external source they wanted to.
Apart from the code of their system,
the students
had to  evaluate their system.

\subsection{Examination of the  Projects}

The delivered software should take as input any JSON file like the one of the collection.
This enables  the  TA (Teaching Assistant) to use a slightly different JSON file during the grading/examination.
Moreover, the students have to deliver a  minimal  User Interface  (e.g. a console interface)
enabling the user of the system  
(a) to select the text  (by selecting one topic id from the JSON file),
(b) to type  a question in natural language,
and 
(c) to view the response of the system.
The format of response is free; apart from the short textual answer 
that is required for computing the metrics, 
students are free to provide responses having more complex form.


\comment{==============
			\subsection{Comparing Projects}
			
			Student's projects can be compared by comparing the measurements
			described in Section \ref{sec:AverageResults}.
			
			A simple metric is: the number of questions are answered correctly.

			The ``big picture" can be illustrated  by filling a table
			like Table \ref{tbl:SummaryOfResultsFromSeveralSystems}.
			

			\begin{table}[!ht!]
			\centering
			\begin{tabular}{|l||l|l|l|l||l|l|l|l||l|}
			\hline
			Group &  \multicolumn{3}{|c|}{Precision}  & \multicolumn{3}{|c|}{Recall}    & \multicolumn{3}{|c|}{F1}  \\\hline
			      &  Max & Min & Avg   &  Max & Min & Avg       &  Max & Min & Avg   \\\hline
			A       &  0 & 0 & 0        &  0 & 0 & 0    &  0 & 0 & 0 \\\hline
			B       &  0 & 0 & 0        &  0 & 0 & 0    &  0 & 0 & 0 \\\hline
			C       &  0 & 0 & 0        &  0 & 0 & 0    &  0 & 0 & 0 \\\hline\hline
			All       &  0 & 0 & 0        &  0 & 0 & 0    &  0 & 0 & 0 \\\hline
			\hline
			\end{tabular}
			\caption{Summarising the Performance of Several (Student's) Systems }
			\label{tbl:SummaryOfResultsFromSeveralSystems}
			\end{table}
			
			\ \\
			Analogously,
			for each metric
			a figure 
			can show 
			the corresponding values that a project 
			achieved 
			(for $n$ projects, that figure will have $n$ functions).
=========}

\subsection{The Projects of Spring 2019}

All projects focused on the first group of topics (1-12).
Each student dedicated around 20 hours.
The accuracy that they achieved ranges 16\%-37\% (i.e. 10 to 23 correctly answered questions from the 61).
The approach that each project followed is summarized below.

\projectDescription{2019-Project1}
{
	Every topic is treated as a separate document corpus and each sentence in the topic's text as a document. 
	Stemming and stopwords removal are applied to every sentence.
	The Okapi BM25 scoring function is used to rank sentences,
	while Stanford CoreNLP with a NER pipeline 
	is used for detecting entities within the text. 
	The "wh-word" 
	as well as  other keywords 
	in the question,
	are used for defining a set of relevant entity types. 
	All entities of a relevant type are sorted 
	based on the BM25 score of their sentence 
	and the top scored entity 
	is the answer that is returned by the system.
}
{CoreNLP}
{20 hours}
{20/61= 32\%} 

\projectDescription{2019-Project2}
{
	This project is based on Named
	Entity Recognition, stemming (Porter's stemmer)
	and Jaccard similarity.
	The entity type of the sought answer is extracted by simple patterns,
	i.e. by checking if the first word of the question 
	is Who, When, Where, etc.
	All keywords in the text whose entity type is the same as 
	the type of the sought answer, are considered as candidate answers. 
	These candidates are scored based on the Jaccard similarity
	between the sentence they occur and the question.
	If a candidate appears in more than one sentence,
	then the maximum score is kept.
	Before calculating the similarity, stopwords are
	removed from the sentences and also stemming is performed.	
}
{CoreNLP}
{20 hours}
{23/61 = 37\%}

\projectDescription{2019-Project3}
{
Given a question in natural language and the searched entity type, 
this project indexes at first the topics of the collection 
(removed stopwords, stemmed terms) and afterwards the sentences of that topic 
in order to find the most relevant one (by calculating the cosine similarity between the query and each topic/sentence). Following, it analyzes the tokens of the retrieved sentence 
(using coreNLP and user-defined regular expressions) 
and returns the questioned entity. 
In general, the application was based on basic linguistic analysis and searching of simple patterns in the given snippets. Furthermore, “Wiktionary”, a semantic dictionary, was used for the recognition of semantic relations. 
In order to premium the primary tokens, 
the ``tf" of the related ones (hyponyms, synonyms etc.) was reduced by 20 per cent.
}
{CoreNLP, Wictionary}
{5 days * 8 hours = 40 hours (approx)}
{10/61 = 16\%. 
}

\projectDescription{2019-Project4}
{	

	This project focused on factoid questions.
 	It uses  ElasticSearch\footnote{https://www.elastic.co/} for building an index out of every topic. 
 	Each sentence is indexed as a document with multiple fields. 
 	Using NER from CoreNLP, 
 	a named-enity field holds the corresponding information that is extracted. 
 	An additional field holds the 	coreference information. 
 	At answer time the best sentence is returned, 
 	using a boolean-multimatch 	ES query, 
 	and a filter is applied for retrieving  only the relevant documents (sentences) 
 	that also contain as a field the same 	named-entity type with the question. 
 	If more than one  answers of the same type are identified, 
 	the system splits sentences into phrases and
	returns the one that is closest to the matching query keywords. 
	}
{ElasticSearch, CoreNLP for NER \& Coref}
{25 hours}
{18/61 = 29\%}

\subsection{The Projects of Spring 2020}

\new{
This year we had projects that  considered topics from all groups.
The approach that each project followed and the accuracy obtained, is summarized below.
}

\projectDescription{2020-1}
{
	In this project we employ BERT: a state-of-the-art NLP transformers model by Google. We made use of a pre-trained BERT model, and fine-tuned it for the task of question answering (QA). We employed Stanford Question Answering Dataset (SQuAD) 2.0 for training and validation. We tried several distinct sets of hyperparameters thought the course. 
	As a result we were able to train a number of different models. Some of them showed higher result for type A questions. 
	Others better handled type B (and some of the type C) questions. For open-domain QA we employed both Google and Wikipedia search. 
	The Wikipedia search consists of searching over wiki pages using the python library \textit{Wikipedia}. Google search was presented by iteratively looking at the first N results of a Google query. For both Wiki and Google search, once the relevant page was retrieved  we fed the content of it to our neural network for find the allegedly correct answer. Our model especially shines for closed-domain QA (type A questions). For different models the result varied from 80\% to 90\%.
}
{ pre-trained BERT, Google and Wikipedia search}
{30-40 hours developing,100+ hours training}
{122/226= 54\%}

\projectDescription{2020-2}
{
This Question Answering system is focused on Answer Type Prediction, and Entity Recognition.
It exploits a combination of unstructured data, from the texts of the evaluation collection, and structured data (RDF) by querying a SPARQL service for several questions.
The pipeline for answering a question consists of 2 main stages: Preprocessing and Answer Extraction.
In the preprocessing stage, the type of the question, the expected answer type, and whether the question is a list question is determined,
and entities in the question are matched with a corresponding URI or label.
For answer type prediction, a set of heuristic rules, and word embeddings are used to match the answer type with the most similar class from the DBpedia ontology.
The answer extraction stage ranks the sentences of the topic by their similarity with the question, and produces a set of answers by finding entities that match
the expected answer type in the topic text or the DBpedia knowledge base. The system found at least one correct answer for 42\% of questions from group A,
35\% of questions from group B, and 27\% from Group C.  
}
{ Core NLP, Word Embeddings, SPARQL }
{50 hours (approx)}
{
Accuracy: 25/226=11\%,
Precision@5: 72/226=32\%,
Candidate Binary Recall: 85/226=38\%.
} 

\projectDescription{2020-3}
{
	The project is based on {\tt Node.js} and built with a client-server-architecture, with the server being responsible for performing the question answering task. The final result uses a node package called question-answering, a module by Huggingface
	\url{https://www.npmjs.com/package/question-answering}. 
	It uses a tokenizer to process the input text and feeds it into a DistilBERT neural network model fine-tuned for question answering
	\url{https://github.com/huggingface/node-question-answering}.		
	Without any manually hardcoded information-extracting rules the accuracy is remarkably high for questions of group A (63\%). 
	However, questions of group B (33\%) and C (18\%) are being answered with a lower accuracy since the project does not support open domain QA or answering confirming questions. The presented measurements of accuracy are calculated by summing the f-scores of each answer, 
	where the f-score of an answer is the F2-measure between expected and provided answer, if it is above 0.65, or 0 otherwise. 
	A lot of time and effort has been spent into designing the very best user experience possible, with a focus on making the demonstration page accessible and understandable for users outside of the computer science domain while providing a suitable representation of data for expert users, especially in the benchmark part.
}
{ Huggingface (question-answering), express}
{40 hours
	\comment{
		Regarding the exact time I didn’t measure it very precisely, but I spent 2 nights (10PM - 6AM) just for designing, implementing and refining the user interface (including benchmark). I had also spent time before for the tf-idf approach and I spent time researching about possible solutions. Of course, just downloading and using the question-answering module in a console application would require only a very limited amount of time.
	}
}
{ 
	101/226= 45\%
	%
}

\projectDescription{2020-4}
{
	This project was implemented in Java and relied on OpenNLP tool.
	The system was able to
	identify the main entities provided by the single question/answer in the json file within the text
	A simple comparison was created between the chosen answer and the correct
	one to evaluate the system which has an accuracy of 22\% in group B of questions
	within the document provided with an accuracy of 13\% in general.
}
{ OpenNLP tool}
{approx. 20 hours}
{Accuracy 22\% in group B of questions	and  13\% in general }

\section{How to Get the Collection}
\label{sec:HowToGet}

The collection is public and can be downloaded from  the Open Data Catalog of ISL\footnote{\url{http://islcatalog.ics.forth.gr/dataset/cs563-qa-an-evaluation-collection-for-question-answering}}. 
It is a JSON file structured as  shown in Figure \ref{fig:example}.
The license of the catalog is:  non-commercial use, attribution.


\comment{=== Se sxolio pros to paron
    \section{Supplementary Material [Optional]}
    \label{sec:SM}
    
    The collection is accompanied with software to ease parsing.
    Specifically it contains ...
    
    [Optional]
    It contains code for computing the following metrics.
====}




\section*{Acknowledgement}

Many thanks to the students of the 2019 class
that contributed 
to this report
including
Giorgos Kadilierakis,
Kostas Manioudakis,
Maria-Evangelia Papadaki
and
Michalis Vardoulakis,
\new{
as well as the students of the 2020 class, 
namely, 
Andrei Kazlouski,
Christos Nikas,
Anthimos Karoutsidis,
and 
Alice di Carlo.
}

\bibliographystyle{plain}
\bibliography{bib/QAbibs}

\appendix

\section{Example of a Topic}
\label{sec:Topic}

An example of a topic is given in Figure \ref{fig:example}.

\begin{figure}[htbp]
	\centering
{\scriptsize
\noindent\makebox[\linewidth]{\rule{\textwidth}{0.4pt}}
\begin{verbatim}
  {
    "id": 7,
    "title": "Tiramisu",
    "text": "Tiramisu (from the Italian language, spelled tiramisu [?tirami?su], 
    meaning 'pick me up' or 'cheer me up') is a coffee-flavoured Italian dessert. 
    It is made of ladyfingers (savoiardi) dipped in coffee, layered with a       
    whipped mixture of eggs, sugar, and mascarpone cheese, flavoured with cocoa.        
    The recipe has been adapted into many varieties of cakes and other desserts.        
    Its origins are often disputed among Italian regions of Veneto and 
    Friuli-Venezia Giulia.   Most accounts of the origin of tiramisu date its 
    invention to the 1960s in the region of Veneto, Italy, at the restaurant 
    'Le Beccherie' in Treviso. Specifically, the dish is claimed to have first been
    created by a confectioner named Roberto Linguanotto, owner of 'Le Beccherie'.  
    Category  Dessert recipes Servings 4-6 Energy 	200 Cal (800 kJ) 
    Time 15 minutes + 1 hour refrigeration 
    Difficulty: 2/5   Ingredients for 6–8 people  4 egg whites  2 egg yolks 100 g 
    (1/2 cup) of sugar  500 g (2 1/2 cups) of mascarpone cheese 4 small coffee 
    cups of espresso coffee 400 g of lady fingers (savoiardi) (or sponge cake)
    unsweetened cocoa powder to sprinkle   over before serving               
    Preparation Make espresso coffee, let it cool a bit.    
    Separate the egg yolks and the whites of two eggs in two bowls.     
    Beat sugar into the egg yolks.     
    Beat the Mascarpone into the sweetened yolks.     
    Add two more egg whites to the other two and whisk  until they form stiff peaks.     
    Fold gently egg whites into Mascarpone mixture.     
    Quickly dip both sides of the ladyfingers in the espresso.     
    Layer soaked ladyfingers and Mascarpone in a  large bowl or pan 
    (start with fingers, finish with mascarpone). Sprinkle
    unsweetened cocoa powder on top just before to serve.  Refrigerate for one hour or two.",
    "qa": [
      {
        "question": "What tiramisu means?",
        "answer": "pick me up / cheer me up",
        "entity": "UNKNOWN",
        "note": "group A, factoid"
      },
      {
        "question": "From which Italian regions has its origins this dessert?",
        "answer": "Veneto and Friuli-Venezia Giulia",
        "entity": "LOCATION",
        "note": "group A, factoid"
      },
      {
        "question": "What is the name of the restaurant that the first tiramisu was made?",
        "answer": "Le Beccherie",
        "entity": "ORGANIZATION",
        "note": "group A, factoid"
      },
      {
        "question": "Who has made the first tiramisu?",
        "answer": "Roberto Linguanotto",
        "entity": "PERSON",
        "note": "group A, factoid"
      },
      {
        "question": "What kind of coffee is used in tiramisu?",
        "answer": "espresso",
        "entity": "INGREDIENT",
        "note": "group A, factoid"
      },
      {
        "question": "List some ingredients of tiramisu?",
        "answer": "eggs, sugar, mascarpone cheese, espresso coffee, cocoa",
        "entity": "INGREDIENT",
        "note": "group A, list"
      }
    ]
  },

\end{verbatim}
\noindent\makebox[\linewidth]{\rule{\textwidth}{0.4pt}}
}
\caption{An example of a topic}
	\label{fig:example}
\end{figure}

\clearpage



\end{document}